# AN EFFECTIVE CONTEXT-BALANCED ADAPTATION APPROACH FOR LONG-TAILED SPEECH RECOGNITION

*Yi-Cheng Wang[1], Li-Ting Pai[1], Bi-Cheng Yan[1], Hsin-Wei Wang[1], Chi-Han Lin[2], Berlin Chen[1]*

[1]National Taiwan Normal University, Taipei, Taiwan
[2]E.SUN Financial Holding Co., Ltd., Taipei, Taiwan
[1]{yichengwang, 61147095s, hsinweiwang, bicheng, berlin}@ntnu.edu.tw
[2]finalspaceman-19590@esunbank.com

## ABSTRACT

End-to-end (E2E) automatic speech recognition (ASR) models have become standard practice for various commercial applications. However, in real-world scenarios, the long-tailed nature of word distribution often leads E2E ASR models to perform well on common words but fall short in recognizing uncommon ones. Recently, the notion of a contextual adapter (CA) was proposed to infuse external knowledge represented by a context word list into E2E ASR models. Although CA can improve recognition performance on rare words, two crucial data imbalance problems remain. First, when using low-frequency words as context words during training, since these words rarely occur in the utterance, CA becomes prone to overfit on attending to the <*no-context*> token due to higher-frequency words not being present in the context list. Second, the long-tailed distribution within the context list itself still causes the model to perform poorly on low-frequency context words. In light of this, we explore in-depth the impact of altering the context list to have words with different frequency distributions on model performance, and meanwhile extend CA with a simple yet effective context-balanced learning objective [1]. A series of experiments conducted on the AISHELL-1 benchmark dataset suggests that using all vocabulary words from the training corpus as the context list and pairing them with our balanced objective yields the best performance, demonstrating a significant reduction in character error rate (CER) by up to 1.21% and a more pronounced 9.44% reduction in the error rate of zero-shot words.

## 1. INTRODUCTION

Thanks to the scalability and streamline nature of end-to-end (E2E) neural models, general-purpose E2E automatic speech recognition (ASR) systems have achieved unprecedented success, significantly outperforming conventional hybrid ASR models on various benchmark tasks [1]. In real-world scenarios, the long-tailed nature of word distribution causes E2E ASR systems to excel with common words but struggle with uncommon ones [2][3][4]. This, however, is detrimental, particularly because uncommon words, such as domain-specific terms (e.g., contact and geo-location names), often play essential roles, and their misrecognition can severely impact downstream tasks like natural language understanding.

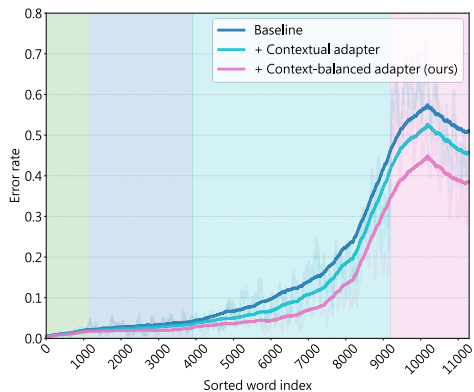

**Figure 1**. Long-tailed problems in AISHELL-1 test-set. (*Many-shot*; *Medium-shot*; *Few-shot*; *Zero-shot*)

In recent years, significant efforts have been made to address this long-tailed issue by incorporating contextual information into E2E ASR systems. This family of methods is known as contextual ASR (CASR) [5]. CASR can be broadly categorized into two approaches: 1) post-training integration of external language models (LM); and 2) integration of contextual information during training. Representative methods of the first category include on-the-fly rescoring with domain-specific *n*-gram or neural language models [6], shallow fusion with a domain-biased weighted finite-state transducer (WFST), and others [7][8][9]. In the second category, the most prevalent methods involve neural contextual adaptation [10][11][12][13][14]. These methods typically encode a list of context words carrying specific contextual information into corresponding embeddings. These embeddings are then infused into the ASR model using a cross-attention mechanism [15] and jointly optimized with the ASR model. Furthermore, the integration of contextual embeddings for context words into the ASR model can be operationalized either in a latent space or in the output distribution [16][17] of the ASR model. Among the various contextual adaptation methods, the contextual adapter (CA) [12] has shown considerable effectiveness in enhancing the performance of recognition on uncommon words. We, however, in this paper, identify two significant data imbalance issues that need to be addressed for CA. The first issue pertains to the context/no-context imbalance problem. When low-frequency words are used as context words during training, the CA model become prone to overfit on attending to <no-context> tokens due to higher-frequency words not being present in the context list. The second issue is the long-tailed distribution within the context list itself, causing the model to perform poorly on low-frequency context words. As shown in Figure 1, CA recognition performance decreases as the word frequency in the training set decreases.

---

[1] The code is available at:
*https://github.com/Amiannn/espnet/tree/context_balanced_adapter*

In light of this, we explore in-depth the impact of altering the context list to have words with different frequency distributions on model performance, and meanwhile extend CA with a simple yet effective context-balanced learning objective for context words. Specifically, we guide the attention mechanism in the adapter by penalizing the attention score of high-frequency context words and promoting attention to rare ones. A series of experiments conducted on the AISHELL-1 benchmark dataset [18] suggests that using all vocabulary words as the context list and pairing them with our balanced objective yields the best performance, validating the effectiveness of our proposed methods.

In summary, our contributions are at least three-fold:
- **Identification of data imbalance problem in contextual adapter**: We identify a previously under-explored issue of data imbalance in the rarity of training context lists and its impact on model performance for CASR models.
- **Balanced learning objective**: We introduce a novel balanced learning objective that penalizes high-frequency context words and promotes rare ones, effectively guiding the attention mechanism within the contextual adapter.
- **Empirical validation**: Through extensive experiments on the AISHELL-1 dataset, we demonstrate that using all vocabulary words as the training context list and pairing them with our balanced learning objective achieve the best performance, validating the effectiveness of our proposed methods.

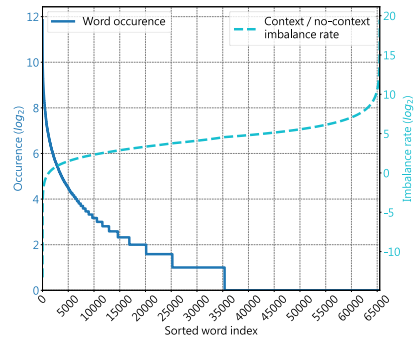

**Figure 2**. Long-tailed word distribution and the context/no-context imbalance rate in the AISHELL-1 training set.

## 2. BACKGROUND

### 2.1. Long-tailed learning

In the scenario of long-tailed learning [19], a small fraction of words have many instances, while most words have only a few instances, as depicted in Figure 2 (blue line). This task is challenging due to two following predicaments: Imbalanced data across words causes ASR models to be biased towards the head words, resulting in poor performance on the tail words. The scarcity of tail words makes it even more challenging to train a robust model that performs well on rare word recognition. To improve the recognition performance of rare words, a contextual adapter (CA) infuses a predefined context list constructed from a set of rarely occurring words into the ASR model. Before delving into the long-tailed problem of CA, we will first introduce some mathematical notions related to long-tailed word recognition, followed by an introduction to our main ASR model. Lastly, we will describe the contextual adapter architecture.

Let $n_w$ be the total number of occurrences of a specific word $w$ in the training set and $N = \sum_w n_w$ denote the total word occurrences in the training set. In long-tailed learning, words are typically sorted by their occurrence $n_w$ in descending order, as shown in Figure 2 (blue line).

### 2.2. Attention-based encoder decoder (AED) model

An AED model generally consists of an encoder network and a decoder network. Given an audio signal O and its corresponding word sequence $W = (w_1, w_2, \cdots, w_U)$. Let $X = (\mathbf{x}_1, \mathbf{x}_2, \cdots, \mathbf{x}_T)$ represent the acoustic feature vectors extracted from O, and $\mathbf{y} = (y_1, y_2, \cdots, y_M)$ be the corresponding character sequence of W. The encoder network, $\text{Enc}^{aco}(\cdot)$, processes a sequence of acoustic feature vectors, X, and produces a high-level representation, $H^{aco} = \text{Enc}^{aco}(X)$, where $H^{aco} \in \mathbb{R}^{T \times D}$. The decoder network, $\text{Dec}(\cdot)$, then fuses the previously decoded text tokens, $\mathbf{y}_{1:m} = (y_1, y_2, \cdots, y_m)$, with the processed acoustic embeddings using a series of cross-attention mechanisms. The output probability of a possible upcoming token $y_{m+1}$ can be derived by

$$P(y_{m+1} \mid X, \mathbf{y}_{1:m}) = \text{Dec}(H^{aco}, \mathbf{y}_{1:m}). \quad (1)$$

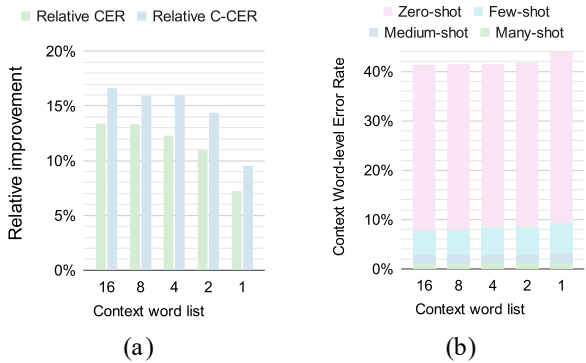

**Figure 3**. (a) The effect of training with different rareness levels of context words, $C^\gamma$, on model performance using the AISHELL-1 test set. The $x$-axis represents the threshold $\log_2(\gamma)$, which is utilized to control the rareness of the context word inside the context list. (b) Long-tailed problems inside contextual adapter, where model perform will on head words (*many-shot*) but fall on tail or unseen words (*zero-shot*) on AISHELL-1 test set.

## 2.3. Contextual adapter (CA)

To improve the recognition accuracy of uncommon or even unseen words during inference, ASR contextual adaptation incorporates a predefined context word list to assist the ASR model in recognizing uncommon words. For instance, in a customer service call center, an ASR system can leverage the caller's account details and previous interactions to provide more accurate recognition.

The contextual adapter (CA) is one of the most effective contextual adaptation methods among recent studies. CA introduces two additional components to the original ASR model: a context encoder, denoted by $\text{Enc}^{ctx}(\cdot)$, and an attention-based adapter. Let $C^\gamma = \{w \mid n_w \leq \gamma\}$ be a context list for training CA, where $\gamma \in \mathbb{Z}^+$ is a threshold that determines the word rarity in the context list and $S$ be the size of $C^\gamma$. The context encoder projects a list of context words into their corresponding context embeddings, represented by $H^{ctx} = \text{Enc}^{ctx}(C^\gamma)$, where $H^{ctx} \in \mathbb{R}^{S \times D}$. The attention-based adapter then integrates these context embeddings into the model through a cross-attention mechanism. Specifically, the CA module is inserted between the encoder network and the decoder network of an AED model. The process of cross-attention and the attention score of the context word $c_s$ at time $t$ can be expressed by

$$\mathbf{q}_t = W^Q \mathbf{h}_t^{aco}, \quad \mathbf{k}_s = W^K \mathbf{h}_s^{ctx}, \quad \mathbf{v}_s = W^V \mathbf{h}_s^{ctx}, \quad (2)$$

$$P(c_s \mid \mathbf{x}_t) = \frac{\exp(\mathbf{q}_t \cdot \mathbf{k}_s / \sqrt{D})}{\sum_{\hat{s}=1}^{S} \exp(\mathbf{q}_t \cdot \mathbf{k}_{\hat{s}} / \sqrt{D})} \quad (3)$$

where $W^Q, W^K$, and $W^V$ represent three distinct learnable linear transformation matrices. Here, $\mathbf{h}_t^{aco} \in H^{aco}$ is the $t$-th acoustic embedding, while, $\mathbf{h}_s^{ctx} \in H^{ctx}$ is the $s$-th context embedding. The attention score $P(c_s \mid \mathbf{x}_t)$ is used to compute a weighted sum of the value embeddings to form the biasing vector: $\mathbf{b}_t^{aco} = \sum_{s \in S} P(c_s \mid \mathbf{x}_t) \mathbf{v}_s$. The acoustic biasing matrix $B^{aco} = (\mathbf{b}_1^{aco}, \mathbf{b}_2^{aco}, \cdots, \mathbf{b}_T^{aco})$ is subsequently utilized to refine the intermediate representations of the acoustic embedding $H^{aco}$ by element-wise addition $\hat{H}^{aco} = H^{aco} + B^{aco}$. Finally, the output probability of a possible next token $y_{m+1}$ can be derived by:

$$P(y_{m+1} \mid X, C^\gamma, \mathbf{y}_{1:m}) = \text{Dec}(\hat{H}^{aco}, \mathbf{y}_{1:m}). \quad (4)$$

### 2.3.1. Training CA

It is worth noting that, the size of the context list, $C^\gamma$, is typically very large (e.g., 10,000). To address this issue during the training phase, random sampling is often employed to create a more tractable subset as a surrogate for $C^\gamma$, denoted by $\hat{C}^\gamma$. This subset $\hat{C}^\gamma$ consists of $\hat{S}$ context words, where $\hat{S} \ll S$. It includes the reference context words $C^{ref}$, context words chosen randomly from the full list $C^\gamma$, and a special token <*no-context*>, which serves as a fallback when there are no relevant context words to consider. Specifically, the reference context list $C^{ref}$ only contain the words that exist both in the reference sentence $W$ and the full context word list $C^\gamma$, which can be describe as $C^{ref} = \{w_u \mid w_u \in C^\gamma \text{ for } u = 1,2,\cdots,U\}$.

The context word subset $\hat{C}^\gamma$ is then utilized to speed up the training of CA. This random context word sampling procedure is closely related to the concept of sampled SoftMax [20].

## 3. DATA IMBALANCE PROBLEMS IN CA

Although the contextual adapter (CA) can enhance the ASR performance on uncommon words, our study has identified two significant data imbalance issues that need to be addressed. The first issue pertains to the imbalance problem of context/no-context words, and the second involves the long-tailed distribution of the context list itself.

## 3.1. Context/no-context imbalance rate

When low-frequency context words are utilized during CA training, most CA studies use named entities or rare words as the context list [12][16][17][21], which typically occur around 16 times or less in the training corpus. This would lead to the CA module overfitting on only attending to the <*no-context*> token. This overfitting happens because higher-frequency or common words occurring in a training utterance are usually absent from the context list, causes the model to inevitably fallback to the <*no-context*> token. This problem is particularly pronounced when the CA module is applied on the acoustic encoder side, as most frames consist of silence.

To characterize this phenomenon, we introduce the notion of the context/no-context imbalance rate, $n_{nctx}/n_{ctx}$, which is the ratio between the total occurrence of context words ($n_{ctx} = \sum_{w \in C^\gamma} n_w$) and the total occurrence of non-context words ($n_{nctx} = N - n_{ctx}$). The context/no-context imbalance rate is illustrated by the cyan line depicted in Figure 2, which indicates that as $\gamma$ becomes smaller, the imbalance rate increases. To observe the effect of different context/no-context imbalance rates on CA performance, we created six different context lists, each containing words of varying rareness. Figure 3(a) shows the CA relative improvements over the baseline AED model across different imbalance rate settings. Surprisingly, when training with a large threshold $\gamma$, the model outperforms those trained with lower thresholds, indicating that lowering the context/no-context imbalance rate is crucial for learning an effective CA module.

### 3.2. Long-tailed problem inside the context list

The long-tailed distribution within the context list, $C^\gamma$, would often make the CA module perform poorly on low-frequency context words, as shown in Figure 3(b). To work around this problem, we extend the CA module with an effective context word balanced training objective, which will be described in the next section.

### 4. CONTEXT-BALANCED ADAPTER

In this section, we introduce our novel context-balanced adaptation approach, aimed at alleviate the imbalance problem of context/no-context words for CA training and the issue of long-tailed distribution within the context word list. Unlike previous studies that primarily used low-frequency words as context words for CA training, our approach utilizes all words in the training corpus as the context word list, effectively reducing the imbalance rate between context/no-context words. To address the long-tailed distribution problem within the context list, we designed an effective context-balance objective. This objective guides the attention mechanism in the adapter by penalizing the attention score of high-frequency context words and promoting attention to rare ones. The context-balance objective can be formulated as follows:

$$\mathcal{L}_{balance} = -\sum_{c_s \in C_{ref}} \omega(c_s) \log P(c_s), \quad (5)$$

$$P(c_s) = \frac{1}{T} \sum_{t=1}^{T} P(c_s | \mathbf{x}_t), \quad (6)$$

where $C_{ref}$ is a set of reference context words, $\omega(c_s)$ is a re-balancing function, and $P(c_s)$ is the context prior probability estimated from the posterior probability, i.e., the attention score from the contextual adapter (c.f. Section 2.3.1). Originally, the re-balance function can simply return the inverse context word frequency. However, recent studies have shown poor performance with this strategy. Instead, the class-balanced loss (CB) [22] introduced a novel concept of effective number to approximate the expected sample number of different classes (in this case, words), which is an exponential function of their training label number. Following this, our re-balance function $\omega(c_s) = (1 - \alpha)/(1 - \alpha^{n_{c_s}})$ is defined to be inversely proportional to the effective number of context words $n_{c_s}$, where $\alpha$ is the hyperparameter controlling the degree of re-weighting. Specifically, $\alpha = 0$ corresponds to no re-weighting, and $\alpha = 1$ corresponds to re-weighting by inverse word frequency.

Apart from our context-balance objective, we also leverage the connectionist temporal classification (CTC) loss to speed up the convergence of the adapter and guide the attention to be aligned with the correct context word at the correct frame, as proposed in [23]. The CTC guidance objective can be derived as follows:

$$\mathcal{L}_{ctc} = -\log \sum_{(c_1, \cdots, c_T) \in \beta^{-1}(C)} \prod_{t=1}^{T} P(c_t = c_s | \mathbf{x}_t), \quad (7)$$

where $\beta^{-1}(\cdot)$ returns all possible alignments compatible with the reference context word sequence $C$, and $c_t \in \hat{C}^\gamma$. The overall objective function of our context-balanced adapter becomes:

$$\mathcal{L} = \lambda_1 \mathcal{L}_{aed} + (1 - \lambda_1)(\lambda_2 \mathcal{L}_{ctc} + (1 - \lambda_2)\mathcal{L}_{balance}), \quad (8)$$

where $\mathcal{L}_{aed}$ is the loss function of the original AED model, and $\lambda_1$ and $\lambda_2$ are the multi task learning hyperparameters.

### 5. EXPERIMENTAL SETUP

#### 5.1. Dataset and evaluation metrics

Our experiments were conducted on AISHELL-1 dataset, a widely-used open-source speech corpus for assessing Chinese ASR systems. AISHELL-1 comprises over 170 hours of Mandarin speech data from various domains, including "Finance," "Science and Technology," "Sports," "Entertainment," and "News."

We assessed our method using the character error rate (CER) and context CER (C-CER). Additionally, we calculated word-level error rates and context word-level error rates for many-shot ($n_w > 100$), medium-shot ($100 \geq n_w > 20$), few-shot ($20 \geq n_w > 0$), and zero-shot ($n_w = 0$) scenarios.

#### 5.2. Baseline and model configuration

We used a Conformer-Transformer model as the backbone for long-tailed speech recognition experiments. The network consists of a Conformer encoder [24] and a Transformer decoder [15] (denoted by Conformer for short). The Conformer encoder consists of 12 blocks, each with 2,048 hidden units and 8 attention heads. The Transformer decoder network includes 6 blocks, each also with 2,048 hidden units.

Table 1. Main results of our context-balanced adapter on the AISHELL-1 test set.

| Model | CER / C-CER | (Word-level / Context Word-level) Error Rate (%) | | | |
|---|---|---|---|---|---|
| | | Many | Medium | Many | Zero |
| Baseline | 4.65 / 8.40 | 0.97 / 1.40 | 2.09 / 2.14 | 7.13 / 7.14 | 37.10 / 36.97 |
| + Contextual Adapter | 4.08 / 7.06 | 0.83 / 1.11 | 2.04 / 1.87 | 5.32 / 5.59 | 33.09 / 32.95 |
| + Context-balanced Adapter (ours) | **3.45 / 5.65** | **0.83 / 1.03** | **1.36 / 1.37** | **3.70 / 3.70** | **27.66 / 27.50** |

Table 2. Impact of the different settings of $\gamma$, $\lambda_2$, and $\alpha$ on the AISHELL-1 test set. It is worth noting that when $\gamma = 2^{16}$ it means that all words from the training corpus were used as the context list for model training.

| $\gamma$ | $\lambda_2$ | $\alpha$ | CER / C-CER | (Word-level / Context Word-level) Error Rate (%) | | | |
|---|---|---|---|---|---|---|---|
| | | | | Many | Medium | Few | Zero |
| $2^2$ | 1 | - | 4.14 / 7.19 | 0.87 / 1.17 | 1.74 / 1.77 | 5.59 / 5.59 | 33.47 / 33.33 |
| $2^4$ | 1 | - | 4.08 / 7.06 | 0.83 / 1.11 | 2.04 / 1.87 | 5.32 / 5.59 | **33.09 / 32.95** |
| $2^8$ | 1 | - | 4.03 / 7.06 | 0.87 / 1.08 | 1.68 / 1.71 | 5.22 / 5.22 | 33.67 / 33.52 |
| $2^{16}$ | 1 | - | **4.03 / 7.00** | **0.79 / 1.03** | **1.68 / 1.71** | **5.06 / 5.06** | 33.71 / 33.57 |
| $2^2$ | 0.5 | 0.9 | 4.09 / 7.12 | 0.85 / 0.99 | 1.76 / 1.62 | 5.57 / 5.23 | 32.78 / 32.80 |
| $2^4$ | 0.5 | 0.9 | 3.97 / 6.76 | 0.83 / **0.95** | 1.62 / 1.51 | 5.26 / 4.98 | 31.51 / 31.56 |
| $2^8$ | 0.5 | 0.9 | 3.83 / 6.59 | 0.86 / 1.15 | 1.60 / 1.61 | 5.02 / 5.02 | 31.08 / 30.93 |
| $2^{16}$ | 0.5 | 0.9 | **3.45 / 5.65** | **0.83** / 1.03 | **1.36 / 1.37** | **3.70 / 3.70** | **27.66 / 27.50** |
| $2^{16}$ | 0.5 | 0.99 | 3.85 / 6.62 | 0.88 / 1.05 | 1.62 / 1.52 | 4.98 / 4.73 | 31.04 / 30.94 |
| $2^{16}$ | 0.5 | 0.999 | 3.83 / 6.55 | 0.87 / 1.14 | 1.62 / 1.64 | 5.13 / 5.13 | 30.70 / 30.55 |
| $2^{16}$ | 0.5 | 0.9999 | 3.82 / 6.52 | 0.87 / 1.15 | 1.64 / 1.66 | 5.08 / 5.08 | 30.68 / 30.53 |

Our proposed method is compared against the iconic contextual adapter (CA) method, which serves as a strong baseline in this study. The context encoder in CA includes an embedding layer and a single-layer BiLSTM with 256 hidden units. The embedding layer is initialized with the corresponding layer from the decoder network. The attention-based adapter comprises a cross-attention module with an embedding size of 256. As for our proposed context-balanced adapter, we adopted the same architecture of context encoder and attention-based adapter configuration as CA. We employed an adapter-style training regime, updating only the context encoder and the attention-based adapter, while keeping the remainder of the ASR model frozen. This approach limits parameter tuning to just 5% of the entire model. The training parameters are set as follows: $(\lambda_1, \lambda_2) = (0.5, 1)$ for the CA method, and $(\lambda_1, \lambda_2) = (0.5, 0.5)$ for the context-balanced adapter. Our proposed CA module was train with the Adam optimizer for 20 epochs with a learning rate of 0.0005.

### 5.3. Context word list configurations

Since Chinese lacks explicit word boundaries, we first perform word segmentation on the AISHELL-1 corpus, and filtering out single-character words (e.g., "我," "他").

**Training phase:** For constructing the context list during model training, we set $\gamma = 2^4$ for the CA method, which is a common setting, and to $\gamma = 2^{16}$ for our context-balanced adapter, which includes all words from the training corpus. The size of the training context list subset $\hat{S}$ is set to 200 for all experiments mentioned above.

**Inference phase:** For constructing the context list for model evaluation, we collected words that occur less than 10 times in the test set to form our testing context list. For all evaluation, we set $\hat{S}$ to 200.

### 6. EXPERIMENTAL RESULTS

### 6.1. Main results

Table 1 presents experimental results on the test set of AISHELL-1 for the baseline ASR model equipped with various contextual adapters (viz., iconic contextual adapter and the proposed context-balanced Adapter). For each method, the CER and C-CER is reported to evaluate the holistic model performance, and the word-level error rate is presented to assess the accuracy of different frequency of words and context words. From Table 1 we can make the following observations. First, for the baseline system, the word-level error rate of the rare words is significantly higher than that of common words. For example, there is a notable performance gap in error rate between many-shot and few-shot testing conditions. This result underscores the long-tailed issues, highlighting that rarely occurring words pose a substantial challenge to existing ASR systems. Second, integrating a list of context words into the Conformer encoder using a contextual adapter enhances the performance of C-CER and across various shots, especially in zero-shot and few-shot scenarios. Lastly, employing all vocabulary words as context words and training the contextual adaptor with the proposed context-balanced objective yields the best overall CER performance.

### 6.2. Detailed analysis

**Impact of different parameter settings on the proposed context-balanced adapter.** The upper part of Table 2 presents an empirical study on the AISHELL-1 test set, examining the impact of various parameter settings of $\gamma$ and $\lambda_2$ values on overall performance of the proposed method, where $\gamma$ controls the rareness of the training context word list, while $\lambda_2$ determines the combination weight of the context-balanced objective.

At outset of experiments, we fix $\lambda_2 = 1$, and probe the impact of the rarity of the training context word list. From this table, we can observe that using high-frequency words as the context list ($\gamma = 2^{16}$) in training phase can improve the performance of the contextual adapter. However, this performance gain is primarily driven by improvements in many-shot and medium-shot words, with limited enhancement observed for unseen words. This underscores that long-tailed distribution problems also exist in the context list. Next, we examine the usage of the context list in the proposed context-balanced adapter by setting $\lambda_2 = 0.5$ with various rareness $\gamma$. We can observe that the combination of using a high-frequency context list $\gamma = 2^{16}$ and incorporating the context-balanced loss $\lambda_2 = 0.5$ yields the best performance in terms of CER, C-CER, and across all shot categories. It is noteworthy that after adding the balanced objective for model training, the recognition performance for many-shot and medium-shot words diminishes slightly, but the performance for few-shot and unseen words improves significantly, as visually displayed in Figure 1.

**Impact of different settings of $\alpha$.** The button part of Table 2 presents an empirical study on the AISHELL-1 test set, examining the impact of different context-balanced hyperparameter $\alpha$ values on overall ASR performance. Following Cui et al. [22], we report various $\alpha$ values {0.9, 0.99, 0.999, 0.9999} in the training phase. It is evident that setting $\alpha$ to 0.9 yields the best performance.

**Qualitative analysis of the attention weights in the contextual adapter.** In Figure 4, we demonstrate the attention weights in the contextual adapter to visualize the impact of training the contextual adapter with the context-balanced objective. Given a test utterance, "因為聚集了過多公共資源," we can observe that the attention weights of contextual adapter tends to densely distribute on the *<no-context>* token. To cope with this issue, integrating the proposed context-balanced loss into the optimization process of contextual adapter

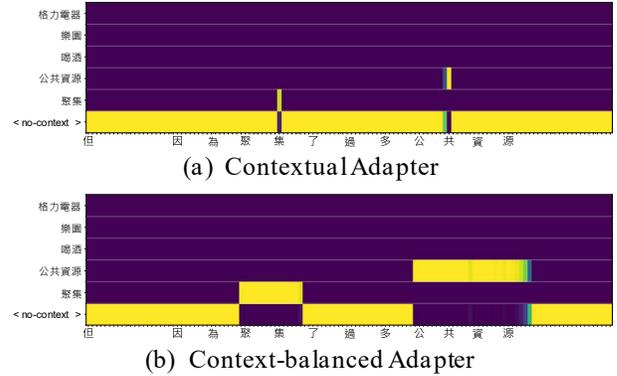

Figure 4. Visualization of the attention scores within the adapter, "聚集" and "公共資源" are the reference context words, $C^{ref}$, in the exemplar utterance. The (a) contextual adapter suffer from overfitting on attending to *<no-context>* token. In contrast, our (b) context-balanced adapter can effectively alleviate the context/no-context imbalance problem.

with can effectively attend to the correct frames when the corresponding context words occur.

### 7. CONCLUSION

In this paper, we have addressed the challenges posed by the long-tailed distribution of words in E2E ASR systems. We identified the issue of data imbalance in training context lists and its impact on ASR performance, particularly for rare words. To mitigate this, we proposed a novel balanced learning objective that penalizes high-frequency context words while promoting attention to rare ones, for enhancing the performance of the contextual adapter (CA) model. Our extensive experiments on the AISHELL-1 benchmark dataset demonstrate that incorporating all vocabulary words as context words to form the context list, combined with our balanced learning objective, significantly improves recognition performance, especially for rare words. These results validate the effectiveness of our proposed methods and highlight the importance of addressing data imbalance in contextual ASR systems. Future work will explore more sophisticated forms of the balanced learning objective and investigate its applicability to other languages and datasets. We believe that our approach provides a promising avenue to improve the recognition of rare words in E2E ASR models, facilitating them to be more reliable and amenable to real-world applications.

### 8. ACKNOWLEDGEMENT

This work was supported in part by E.SUN Bank under Grant Numbers 202308-NTU-02 and 202408-NTU-02. Any findings and implications in the paper do not necessarily reflect those of the sponsor.